\documentclass[conference]{IEEEtran}
\usepackage{cite}
\usepackage{graphicx}
\usepackage{amsmath,amssymb,amsfonts}
\usepackage{booktabs}
\usepackage{tabularx}
\usepackage[caption=false,font=footnotesize]{subfig} % IEEE recommended for subfigures

\usepackage{listings}
\usepackage{textcomp}
\usepackage{xcolor}

\usepackage{listings}
\usepackage[linesnumbered,ruled,vlined]{algorithm2e}
\SetAlFnt{\footnotesize}
\SetAlgoCaptionLayout{centerline}

\usepackage[hidelinks]{hyperref}

\def\BibTeX{{\rm B\kern-.05em{\sc i\kern-.025em b}\kern-.08em
  T\kern-.1667em\lower.7ex\hbox{E}\kern-.125emX}}

\lstdefinelanguage{yaml}{
  keywords={true,false,null,y,n},
  keywordstyle=\bfseries\color{blue},
  basicstyle=\ttfamily\scriptsize,
  sensitive=false,
  comment=[l]{\#},
  morecomment=[s]{/*}{*/},
  commentstyle=\color{gray}\ttfamily,
  stringstyle=\color{red}\ttfamily,
  moredelim=[l][\color{orange}]{\&},
  moredelim=[l][\color{magenta}]{*},
  moredelim=**[il][\color{black}:]{:},
  morestring=[b]', morestring=[b]"
}
\title{Language Model Guided Reinforcement Learning in Quantitative Trading}

\author{
  \IEEEauthorblockN{Adam Darmanin}
  \IEEEauthorblockA{University of Malta\\Msida, Malta\\ adam.darmanin.03@um.edu.mt}
  \and
  \IEEEauthorblockN{Vince Vella}
  \IEEEauthorblockA{University of Malta\\Msida, Malta\\ vvell04@um.edu.mt}
}

\begin{document}

% Invoke IEEEtran BST control entry (defined in references.bib) before first citation
% so that style controls (e.g., dashed repeated names, URL spacing) take effect.
\bstctlcite{IEEEexample:BSTcontrol}

\maketitle

\begin{abstract}
Algorithmic trading requires short-term tactical decisions consistent with long-term financial objectives. Reinforcement Learning (RL) has been applied to such problems, but adoption is limited by myopic behaviour and opaque policies. Large Language Models (LLMs) offer complementary strategic reasoning and multi-modal signal interpretation when guided by well-structured prompts.

This paper proposes a hybrid framework in which LLMs generate high-level trading strategies to guide RL agents. We evaluate (i) the economic rationale of LLM-generated strategies through expert review, and (ii) the performance of LLM-guided agents against unguided RL baselines using Sharpe Ratio (SR) and Maximum Drawdown (MDD).

Empirical results indicate that LLM guidance improves both return and risk metrics relative to standard RL.
\end{abstract}


\section{Introduction}
Algorithmic trading requires short-term execution aligned with long-term financial objectives, and accounts for over 60\% of U.S. equity volume, particularly in high-frequency trading (HFT)~\cite{Chlistalla2011HFT}. Long-horizon strategies often build on econometric models such as the Fama--French five-factor framework~\cite{FamaFrench2015}, while short-horizon approaches exploit transient inefficiencies through momentum and mean reversion~\cite{Theate2021}. The most extreme form is HFT, where firms execute thousands of transactions per second by exploiting order book imbalances~\cite{Chlistalla2011HFT}.

Modern trading systems leverage machine learning (ML) to process structured and unstructured data streams. Growth is driven by advances in electronic infrastructure, compute power, and the proliferation of large high-resolution financial data~\cite{bartram2020artificial}.

RL formalises trading as sequential decision making~\cite{Sutton1998} and has shown promise through methods such as Deep Q-Networks (DQN) and actor--critic variants~\cite{Theate2021,Liu2021FinRL}. Yet practical adoption is hindered by sparse rewards, credit assignment issues, and opaque policies, which reduce trust in high-stakes financial settings~\cite{LopezdePrado2020BeyondEconometrics}.

LLMs offer complementary strengths. They can understand heterogeneous signals and generate rationale explanations~\cite{Onozo2024FinNews}, yet remain limited by fixed knowledge cut-offs and an inability to adapt to changing environments in real time~\cite{zhang2024multimodalfoundationagentfinancial}. They are also fragile to prompt design and may produce plausible but invalid outputs~\cite{wang2024llmfactorextractingprofitablefactors,schulhoff2024promptreportsystematicsurvey}. Advances in prompt engineering and human-in-the-loop (HITL) methods partially mitigate these risks~\cite{Yu2024FinMem}.

This paper addresses these limitations by proposing a hybrid architecture that integrates LLM strategic guidance with RL execution, and sets the following objectives.  

\paragraph{Objectives} (1) Design an approach for using LLMs to generate trading strategies that are economically grounded, assessed by expert review; (2) evaluate whether LLM guidance improves RL performance, measured by SR and MDD, without altering the base RL architecture.  

These objectives directly inform the main contributions of the paper.  

\paragraph{Contributions} (1) A structured prompting framework that produces domain-grounded strategies assessed by expert review and financial metrics; (2) a modular LLM+RL design in which the LLM contributes a single uncertainty-weighted scalar appended to the RL observation space, improving out-of-sample performance without modifying the RL algorithm.    

\subsection{Related Work}

A Trading DQN (TDQN) for stock trading is introduced in ~\cite{Theate2021}. The authors leverage a DDQN algorithm to mitigate overestimation bias and stabilize learning in stochastic market environments. A key aspect of their RL approach was the discretization of actions and the enforcement of capital constraints, which helped prevent infeasible or overleveraged actions.

The FinRL framework~\cite{Liu2021FinRL} introduced benchmark environments and unified APIs for financial RL research that features realistic data simulation. It includes a wide array of backtests using standard RL algorithms and focuses on two primary objectives in algorithmic trading: maximizing return (measured by cumulative return and SR) and minimizing risk (measured by MDD and return variance). The framework supports experiments in single-stock trading, multi-stock trading, and portfolio allocation.

In the survey~\cite{Arulkumaran2017}, the authors identified key limitations in deep reinforcement learning (DRL), including Bellman backup instability and credit assignment failures. The authors recommend hierarchical reinforcement learning (HRL) or recurrent extensions to address the lack of long-range temporal dependencies.

For LLMs in finance, \cite{lopezlira2023chatgptforecaststockprice} showed that ChatGPT outperforms traditional sentiment lexicons on forward-looking financial news, but lacks temporal awareness and numerical reasoning capabilities.

The FINMEM framework in~\cite{Yu2024FinMem} combines structured memory with LLM-based decision modules. FINMEM’s layered memory integrates recent news, financial reports, and long-term statements to inform trade recommendations, leveraging retrieval-augmented generation (RAG). Their architecture stores experiences in a vector database, which are retrieved and ranked using a decay mechanism that emulates a human's memory decay.

Prompting practices have been extensively surveyed in \cite{schulhoff2024promptreportsystematicsurvey}, which categorizes strategies into instruction-based, example-based, reasoning-based, and critique-based families. The study highlights self-refinement and constraint enforcement as key mechanisms for improving robustness. It also shows that minor variations in prompt wording can systematically influence model behavior.

In \cite{huang2022reasoninglargelanguagemodels}, Chain-of-Thought (CoT) prompting was shown to significantly enhance LLM reasoning. Self-improvement frameworks iteratively refine rationale quality, while problem decomposition and model fine-tuning help address complex tasks. Without structured prompting techniques, however, LLMs continue to struggle with planning problems.

\section{Methodology}

This section outlines the methodology developed to evaluate the integration of LLMs into RL agents. The proposed hybrid framework mirrors the top-down decision-making structures common in financial institutions.

Two experiments were conducted to address the research objectives. All LLM strategies were validated through historical backtesting and expert review prior to their integration with RL agents.

\subsection{Benchmark Environment}

For our benchmark, we utilized the trading system introduced in~\cite{Theate2021}. This benchmark includes a clearly defined environment, consistent state and reward functions, and extensive empirical results. 

We replicated the core experimental settings, including the asset universe, data preprocessing, and evaluation metrics. Our reproduction yielded comparable statistically significant SR and MDD metrics.

We note minor discrepancies that affect financial interpretability: their cumulative returns are arithmetically summed rather than geometrically compounded, and their SR assumes a zero risk-free rate. For consistency, we preserve these conventions throughout our experiments.

\subsection{Experiment 1: LLM Trading Strategy Generation}

\begin{figure*}[t]
  \centering
  % Omit file extension so pdf/eps/png fallbacks work; ensure vector PDF present for IEEE.
  \includegraphics[width=\textwidth]{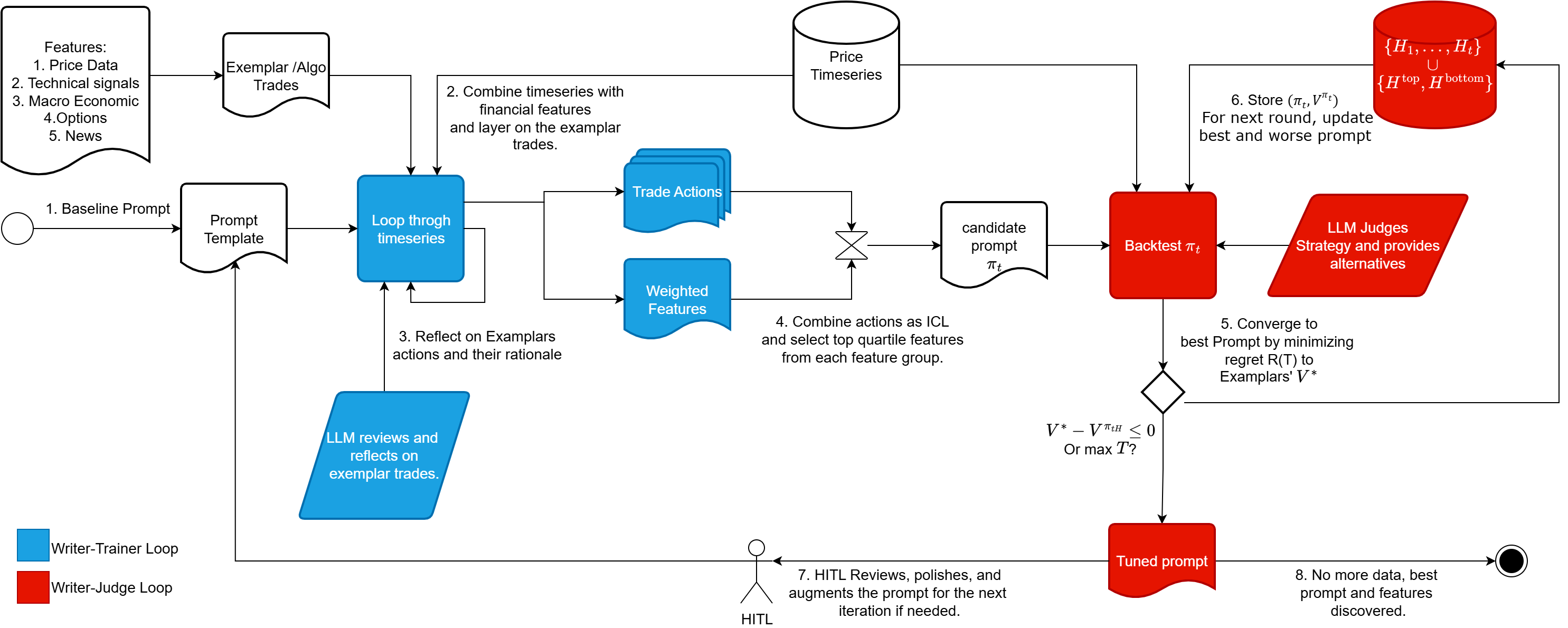}
  \caption{Prompt Tuning Workflow.}
  \label{fig:prompt-tuning}
\end{figure*}

This experiment addressed Objective~1 by introducing two LLM agents: the \textbf{Strategist Agent} and the \textbf{Analyst Agent}. The Strategist Agent generates global trading policies using a financial dataset. The Analyst Agent processes news and distills it into signals to inform the Strategist Agent. This experiment serves as the foundation for Experiment~2.

A strategy defines a directional action ($\text{dir}(\pi^g)$, where 1 = \texttt{LONG} and 0 = \texttt{SHORT}) and an associated confidence score ($\mu_{\text{conf}}$, from 1 to 3). Each strategy is accompanied by an explanation and a weighted set of features. Strategies are generated on a monthly basis using time-aligned data.

\subsubsection{Data and Feature Engineering}
\label{sec:llm_method_data}

To support strategy generation, the LLM agents consumed a multi-modal dataset spanning 2012–2020, aligning with the benchmark's dataset dates in \cite{Theate2021}. The dataset includes traditional Open, High, Low, Close, and Volume (OHLCV) price data, which we augmented with four additional categories of financial signals: market data, fundamentals, technical analytics, and alternative data~\cite{LopezdePrado2018ML}. These collectively define the LLM's context.

Market data was sourced from Interactive Brokers\footnote{\url{https://www.interactivebrokers.com/api}} and iVolatility\footnote{\url{https://www.ivolatility.com/data-cloud-api/}}, including OHLCV time series, SPX and NDX index returns, the VIX index, and Options implied volatility (IV). Fundamental data, comprising firm-level financial ratios and macroeconomic indicators (e.g., GDP, PMI, interest rates), were retrieved via SEC-API\footnote{\url{https://sec-api.io/}} and the FRED API\footnote{\url{https://fred.stlouisfed.org/docs/api/fred/}}. Analytics features were computed using TA-Lib\footnote{\url{https://ta-lib.org/}}, applying rolling windows to extract indicators. Alternative data consisting of news headlines were collected from Alpaca\footnote{\url{https://alpaca.markets/docs/api-documentation/}} and processed into explanatory factors using few-shot LLM prompting, following the LLMFactor framework from~\cite{wang2024llmfactorextractingprofitablefactors}. The data was aligned by timestamp.

\subsubsection{LLM Model}
\label{sec:mat_llmbackbone}

We used OpenAI’s GPT-4o Mini for its strong performance in financial reasoning and cost-efficiency~\cite{lopezlira2023chatgptforecaststockprice, Yu2024FinMem, huang2022reasoninglargelanguagemodels}. The model supports a 128k token context window with a 16k maximum prompt size, enabling the use of detailed prompts with embedded context memory, reasoning chains, and previous reflection results.

\subsubsection{Prompt Engineering Methodology}
\label{sec:prompt_gen}

The objective was to construct a prompt that generalized across equities and regimes while remaining interpretable. 
We proceeded in three stages: (i) baseline specification, (ii) incorporation of expert exemplars with feature pruning, and (iii) iterative refinement comprising two distinct processes: a \emph{Writer--Trainer} process (feature and instruction selection) and a \emph{Writer--Judge} process (prompt quality and rationale critique) with regret minimization.

To manage computational cost, we did not tune on full eight-year history and instead randomly sampled non-overlapping one-year intervals from the dataset per instrument, repeated five times, and used these subsets for creating candidate prompts.

\paragraph{Baseline}
We began with a minimal prompt that exposed only raw OHLCV data and a small set of technical indicators: Simple Moving Averages (SMA; 20/50/200 periods), Relative Strength Index (RSI), and Moving Average Convergence Divergence (MACD). This configuration reflected common trading heuristics in both algorithmic and retail practice~\cite{Lucas2024DQNSparse,chaddha2022predictive} and served to set the candidate prompt and \(V^*\) target.

\paragraph{Writer--Trainer, and Writer--Judge Loops}
To refine the information set, we introduced a \emph{Writer--Trainer} process that reflected on expert trade exemplars derived from HITL feedback and a heuristic algorithm approximating these. Candidate features were selected by top quartile through ranking importance on a three-point Likert scale (low/medium/high), and rationales were clustered with only the ten unique ones being selected to become instructions for the candidate prompt. 

Prompt refinement was then conducted through a heuristic regret–minimization loop, inspired by \cite{wang2024quantagentseekingholygrail}, with pruning and rationale discovery as the inner stage and backtesting as the outer stage. At each iteration $t$, a \emph{writer} generated a candidate prompt $\pi_t$ conditioned on the KB and the retained features and rationales. The candidate $\pi_t$ was evaluated through backtesting to obtain its SR, denoted $V^{\pi_t}$. A \emph{judge} then assessed the prompt for its rationale and suggested alternative instructions or feature combinations for the next iteration, and stored in the Knowledge Base (KB)

We adopted a regret heuristic to guide exploration:
\begin{equation}
\mathcal{R}(T) \;=\; \mathbb{E}\!\left[\sum_{t=1}^T \bigl(V^* - V^{\pi_t}\bigr) \,\middle|\, \mathcal{H}_t \right],
\label{eq:regret}
\end{equation}
where $V^*$ denotes the best SR defined by the baseline prompt or the market's SR for the whole dataset (initialized as $V^*=\max\{V_{\text{baseline}},\,0.8\}$), $V^{\pi_t}$ is the SR of the current candidate, and $\mathcal{H}_t$ represents the KB at iteration $t$ (features, rationales, and prior outcomes). To manage the LLM’s finite context window, the loop retained only an \emph{extremes memory} within $\mathcal{H}_t$, consisting of the best- and worst-performing prompts with their associated features and instructions. These extremes biased subsequent generations toward more promising candidates.

Iterations terminated when either (i) $\mathcal{R}(T)\leq 0$ (no expected improvement) or (ii) the maximum $T$ iterations elapsed. The final augmented baseline prompt, generalized across equities and regimes through this process, was then subjected to three additional refinements.

\paragraph{Prompt Improvement 1 -- In-Context Memory (ICM)}  
Inspired by~\cite{Yu2024FinMem}, we introduced a memory buffer that stores the most recent strategies $\pi_t$ observed prior to time~$T$. Each stored strategy $\pi^{g}_{T-1}$ is represented by its directional action, weighted features, and rationale. Within the prompt, these prior strategies are recalled in-context and reflected on, enabling the current strategy $\pi^{g}_{T}$ to be conditioned on past decisions. This reflection mechanism reduces the persistence of suboptimal strategies.

\paragraph{Prompt Improvement 2 -- Instruction Decomposition} 
To enhance reasoning, instructions and their associated drivers were decomposed~\cite{schulhoff2024promptreportsystematicsurvey} into six feature groups: stock, technical, fundamental, macroeconomic, options, and prior strategy reflection~\cite{zhang2024multimodalfoundationagentfinancial}. Each group supplied few-shot examples and domain-specific heuristics to elicit CoT, requiring the model to reason sequentially across domains and prior strategies.

\paragraph{Prompt Improvement 3 -- News Factors} Unstructured news data was introduced via the analyst agent, which applied instruction-decomposition factor extraction~\cite{wang2024llmfactorextractingprofitablefactors}. Entities and timestamps were anonymized, disabling the LLM’s memory to prevent leakage~\cite{lopezlira2025memorizationproblemtrustllms}. Extracted news factors were ranked and integrated alongside numerical indicators. 

The final system integrates selected numerical and textual signals into a global strategy policy \( \pi^g \). All prompt iterations used in Experiment~1 are summarized in Table~\ref{tab:prompt_definitions}.

\begin{table}[t]
\centering
\footnotesize
\caption{Prompt Versions Used in Experiment 1}
\label{tab:prompt_definitions}
\begin{tabular}{@{}l p{6.9cm}@{}}
\toprule
\textbf{Prompt} & \textbf{Description} \\
\midrule
P0 & Baseline prompt containing only static technical indicators and price features. \\
P1 & Augmented P0 with selected features and instructions. \\
P2 & P1 extended with ICM, incorporating prior strategy. \\
P3 & P2 extended with instruction decomposition and CoT reasoning across six structured signal groups. \\
P4 & P3 enriched with macroeconomic and firm-specific news-derived directional signal. \\
\bottomrule
\end{tabular}
\end{table}

\subsubsection{Parameters and Evaluation}
\label{sec:llm_param}

Prompt tuning used temperature $0.7$ following prior work~\cite{Yu2024FinMem,lopezlira2023chatgptforecaststockprice}, with frequency penalty $1.0$ and presence penalty $0.25$. These values discouraged verbatim reuse from the KB or ICM while preserving exploratory diversity.

For strategy generation, temperature was set to $0$ with fixed seed $49$ for reproducibility. Strategies were produced on a monthly cadence (20 trading days), aligning with common guidance/rebalancing cycles and remaining tractable given LLM inference cost.

At most three refinement iterations were permitted ($T\le 3$). Convergence was declared when the regret $\mathcal{R}(T)$ approached zero or when the SR exceeded the initial threshold $\max\{V_{\text{baseline}},\,0.8\}$. The procedure was repeated five times with discretionary HITL adjustments between runs. The iteration count balanced methodological tractability against computational cost.

All technical indicators used a 20-trading-day rolling window with standard \textit{TA-Lib} defaults (e.g., 14-day RSI).

\paragraph{Quantitative Metrics} We evaluate LLM-generated strategies using three complementary metrics: risk-adjusted returns, model confidence, and model uncertainty.

The SR serves as the core risk-adjusted returns metric:
\begin{equation}
\text{SR} = \frac{\mathbb{E}[R_t - R_f]}{\sigma_R}.
\label{eq:sr}
\end{equation}
where \( R_t \) is the portfolio return, \( R_f \) is the risk-free rate, and \( \sigma_R \) is the return volatility. SR also serves as a proxy for the LLM’s financial reasoning~\cite{lopezlira2023chatgptforecaststockprice, Yu2024FinMem}. To ensure comparability across different periods for the daily returns, we annualize the SR to 252 trading days per year: \( \text{Annualized SR} = \text{SR} \cdot \sqrt{252} \).

As a proxy for prompt quality we compute the Perplexity (PPL)~\cite{gonen-etal-2023-demystifying} over the LLM-generated strategies:
\begin{equation}
\text{PPL} = \exp\left( -\frac{1}{N} \sum_{t=1}^{N} \log p(w_t \mid w_{<t}) \right).
\end{equation}
where \( p(w_t \mid w_{<t}) \) denotes the conditional token probability. Lower values indicate higher quality.

To complement this, we report token-level entropy \( H_{\text{LLM}} \), approximated using top-\(k\) distributions:
\begin{equation}
H_{\text{LLM}} = \frac{1}{N} \sum_{t=1}^N \left( \sum_{v \in V_k} -p_t(v) \log p_t(v) - p_{\text{tail},t} \log p_{\text{tail},t} \right).
\label{eq:trunc-ent}
\end{equation}
where \( V_k \) denotes the top-\(k\) token set and \( p_{\text{tail},t} \) represents the unobserved probability mass~\cite{kaltchenko2025entropy}. In our experiments, \(k=5\). Lower entropy indicates greater decisiveness, whereas higher values suggest uncertainty.

Together, PPL and \( H_{\text{LLM}} \) enable a measurement of prompt quality and strategy confidence.

\label{sec:exp1_qualitative}
\paragraph{Qualitative Evaluation} Qualitative assessment was conducted via an Expert Review Score (ERS), a human-grounded rubric evaluating LLM-generated trading rationales along three dimensions: economic rationale, domain fidelity, and trade safety (risk awareness). Each dimension was scored on a 3-point ordinal scale \{1 = poor, 2 = average, 3 = good\}, based on the rubric shown in Table~\ref{tab:expert_rubric}.

The review process followed a similar setup to that of~\cite{Demajo_2020}, involving ten participants: five senior finance professionals and five retail traders (or professionals in the industry who do not actively trade). Each reviewer evaluated anonymized data for three instruments over one year, including price data, fundamental and macroeconomic metrics, and firm-level news headlines.

Before reviewing the LLM rationale, expert participants made their own directional prediction (\texttt{LONG}/\texttt{SHORT}) to activate their internal domain models. They then reviewed the LLM's reasoning and scored it using the rubric. Each session concluded with a 60-minute structured discussion to elicit LLM critiques and identify exemplars to use. Surveys took approximately 15 minutes to complete. All scores were normalized to a 1–3 range.

\begin{table}[!t]
\centering
\footnotesize
\caption{Expert rubric for scoring LLM rationales}
\label{tab:expert_rubric}
\begin{tabular}{lccc}
\toprule
\textbf{Criterion} & \textbf{1} & \textbf{2} & \textbf{3} \\
\midrule
Rationale & Flawed & Partial & Sound \\
Fidelity  & Unrealistic & Plausible & Professional \\
Safety    & Ignored & Mentioned & Addressed \\
\bottomrule
\end{tabular}
\end{table}

\subsection{Experiment 2: LLM-Guided RL}

This experiment addressed Objective~2 by incorporating the LLM guidance within an RL framework.

\subsubsection{Data and Feature Engineering} The LLM outputs from Experiment~1 were reused. The RL agent adopted the DDQN configuration of~\cite{Theate2021}, with a single LLM-derived interaction term $\tau$ as innovation to the observation space. This feature consisted of:

\begin{itemize}
    \item \textbf{Signal Direction} ($\mathrm{dir}(\pi^g)$): The discrete directional recommendation from the LLM. Zero represents \texttt{SHORT} and one \texttt{LONG}.
    \item \textbf{Signal Strength} ($\mathrm{str}(\pi^g)$): The LLM's entropy-adjusted confidence score as a Likert-3 score.
\end{itemize}

The interaction term was defined as
\begin{equation}
	\tau = \mathrm{dir}(\pi^g) \cdot \mathrm{str}(\pi^g),
\end{equation}
where \(\mathrm{dir}(\pi^g)\) was remapped from \(\{0,1\}\) to \(\{-1,1\}\) to enable the interaction.

The LLM's signal strength was derived from the normalized LLM's confidence score:
\begin{equation}
\mu_{\text{conf}} = \frac{\text{Likert}}{3},
\end{equation}

and adjusted using entropy-based certainty:
\begin{equation}
C = \varepsilon + (1 - \varepsilon)(1 - H),
\end{equation}

where $H \in [0,1]$ is the normalized entropy of the LLM output, and $\varepsilon = 0.01$ ensures numerical stability. The final strength term is:
\begin{equation}
\mathrm{str}(\pi^g) = \mu_{\text{conf}} \cdot C.
\end{equation}

This entropy-adjusted confidence follows the approach of~\cite{yona2024faithful}, providing a soft weighting of the LLM's signal by its certainty.

The interaction term $\tau$  was selected empirically. Initial variants used direction only ($\mathrm{str}(\textit{dir})$), followed by LLM's confidence ($\mathrm{str}(\pi^g)$) and direction. The final form was chosen based on empirical performance and compatibility with DDQN’s continuous normalized input space~\cite{Theate2021}.

\subsubsection{LLM+RL Hybrid Architecture} The baseline DDQN agent is augmented by the Strategist Agent and Analyst Agent, which produce monthly strategies for the stock's behavior. For practical reasons, outputs from the LLM were precomputed per instrument and fixed throughout training.

\subsubsection{Training and Parameters} Hyperparameters mirror \cite{Theate2021} and the LLM settings follow those in Experiment~1. Training was conducted over 25 runs × 50 episodes per instrument using an NVIDIA RTX 3050, with each equity trained for 3 hours.

To ensure comparability with the benchmark~\cite{Theate2021}, we replicated all baseline metrics within acceptable statistical bounds.

\subsubsection{Evaluation Metrics} Two measures were considered:

\begin{itemize}
    \item \textbf{SR:} Same as Experiment~1 see Eq.~\eqref{eq:sr}.
    
    \item \textbf{MDD:} Captures the largest observed loss from a historical peak to a subsequent trough:
  \begin{equation}
    	\text{MDD} = \frac{P_{\text{peak}} - P_{\text{low}}}{P_{\text{peak}}}.
  \end{equation}
    where \( P_{\text{peak}} \) is the highest portfolio value observed before the largest drop, and \( P_{\text{low}} \) is the lowest value reached before a new peak is established. Lower values indicate stronger downside protection.
\end{itemize}

These metrics together assess whether LLM-guided RL agents can adapt to different equities without changing the core architecture.

\section{Results and Discussion}

\subsection{Experiment 1 Results}
\label{sec:exp1_results}

This section presents empirical results across the baseline (P0) and four prompt versions (P1–P4) from Table~\ref{tab:prompt_definitions}, addressing Objective~1. We evaluated their impact on SR, PPL, and $H_\text{LLM}$. From P4 onward, qualitative evaluation was incorporated through ERS, introduced once the prompt design had stabilized. All backtests were conducted over 2018–2020 to ensure comparability with the RL's OOS results. Statistical significance was assessed using two-tailed \textit{t}-tests across 25 runs per ticker, with hypotheses $H_0: \mu_{\text{P4}} = \mu_{\text{P1}}$. All runs were executed at a sampling temperature of 0.7 to capture variance, while the reported metrics correspond to the deterministic setting (temperature 0) with fixed random seeds for reproducibility.

\begin{table}[t]
\centering
\footnotesize
\caption{Sharpe Ratio Across Prompts and Benchmark}
\label{tab:sr_combined_benchmark}
\begin{tabular}{lrrrrrr}
\toprule
\textbf{Ticker} & P0 & P1 & P2 & P3 & P4 & \textbf{BM} \\
\midrule
AAPL  & 1.13 & 1.09 & 1.07 & 1.07 & \textbf{2.09} & 1.27 \\
AMZN  & 0.51 & 0.35 & 0.38 & 0.63 & \textbf{0.84} & 0.21 \\
GOOGL & 0.34 & 0.26 & 0.52 & 0.52 & \textbf{1.12} & 0.19 \\
META  & 0.60 & -0.06 & -0.28 & 0.30 & \textbf{0.77} & 0.63 \\
MSFT  & 0.36 & 1.07 & 1.11 & \textbf{1.31} & 0.50 & 1.17 \\
TSLA  & 0.34 & 0.71 & 0.75 & 0.43 & \textbf{0.79} & 0.67 \\
\midrule
Mean  & 0.55 & 0.57 & 0.59 & 0.71 &\textbf{1.02} & 0.69 \\
\bottomrule
\end{tabular}
\end{table}

\begin{table}[t]
\centering
\footnotesize
\caption{Perplexity Across Prompts}
\label{tab:ppl_combined}
\begin{tabular}{lrrrrr}
\toprule
\textbf{Ticker} & P0 & P1 & P2 & P3 & P4 \\
\midrule
AAPL  & 1.44 & 1.85 & \textbf{1.31} & 1.55 & 1.44 \\
AMZN  & 1.51 & 1.74 & 1.35 & 1.68 & \textbf{1.31} \\
GOOGL & 1.56 & 1.77 & 1.49 & 1.78 & \textbf{1.33} \\
META  & 1.47 & 1.73 & \textbf{1.31} & 1.39 & 1.38 \\
MSFT  & 1.43 & 1.83 & 1.44 & 1.49 & \textbf{1.24} \\
TSLA  & 1.46 & 1.77 & 1.50 & 1.63 & \textbf{1.39} \\
\midrule
Mean  & 1.48 & 1.78 & 1.40 & 1.59 & \textbf{1.35} \\
\bottomrule
\end{tabular}
\end{table}

\begin{table}[t]
\centering
\footnotesize
\caption{Entropy Across Prompts}
\label{tab:entropy_combined}
\begin{tabular}{lrrrrr}
\toprule
\textbf{Ticker} & P0  & P1 & P2 & P3 & P4 \\
\midrule
AAPL  & 0.66 & 0.70 & 0.67 & \textbf{0.66} & 0.69 \\
AMZN  & 0.69 & 0.69 & 0.69 & 0.69 & \textbf{0.67} \\
GOOGL & 0.67 & 0.67 & 0.67 & 0.70 & \textbf{0.66} \\
META  & 0.68 & \textbf{0.66} & 0.70 & 0.73 & 0.67 \\
MSFT  & 0.65 & 0.66 & 0.68 & 0.72 & \textbf{0.65} \\
TSLA  & 0.67 & 0.68 & 0.70 & 0.74 & \textbf{0.65} \\
\midrule
Mean  & 0.67 & 0.68 & 0.69 & 0.71 & \textbf{0.67} \\
\bottomrule
\end{tabular}
\end{table}

\begin{table}[t]
\centering
\footnotesize
\caption{Expert Reviewer Scores for Prompt 4}
\label{tab:prompt4_evaluation}
\begin{tabular}{lr}
\toprule
\textbf{Dimension} & \textbf{ERS (1–3)} \\
\midrule
Rationale & 2.70 \\
Fidelity  & 2.65 \\
Safety    & 2.80 \\
\bottomrule
\end{tabular}
\end{table}

\begin{table}[t]
\centering
\footnotesize
\caption{P4 vs.\ P1 Significance of Metric Changes}
\label{tab:significance_summary}
\begin{tabular}{lc}
\toprule
\textbf{Metric} & \textbf{\textit{t}-test $p$-value} \\
\midrule
Entropy & \textbf{$2.29\times10^{-4}$} \\
Perplexity & $7.25\times10^{-2}$ \\
Sharpe Ratio & \textbf{$2.3\times10^{-5}$} \\
\bottomrule
\end{tabular}
\end{table}

Tables~\ref{tab:sr_combined_benchmark}–\ref{tab:entropy_combined} summarize the results across prompt versions relative to the benchmark (BM). Prompt~0, which relied solely on static technical features, outperformed Prompt~1 primarily because all equities exhibited upward trends during the OOS period. Prompt~1 yielded the weakest performance, with the lowest SR across most equities and the highest PPL and entropy, indicating that the LLM was unable to exploit the additional information when presented in an isolated context. Prompt~2 incorporated ICM, producing moderate gains in SR (mean $0.59$) and suggesting improved confidence through reflection. Prompt~3 introduced decomposed instructions, eliciting CoT, and outperformed the benchmark with a mean SR of $0.71$. Prompt~4 further included unstructured news signals and achieved the highest mean SR ($1.02$), lowest PPL and entropy, and showed higher confidence particularly on sentiment-sensitive tickers such as TSLA. Based on the $p$-values in Table~\ref{tab:significance_summary}, the improvements were statistically significant for SR and entropy, while the changes in PPL were comparatively weaker.

Expert evaluation of Prompt~4 confirmed its effectiveness. Reviewers rated the LLM's rationale highly (mean $2.7$ out of $3$), highlighting its ability to integrate valuation, sentiment, and analytics.

Fidelity received a slightly lower score (mean $2.65$), with critiques focused on inconsistent thresholding. For instance, one reviewer noted, \textit{“Calling RSI near $40$ ‘oversold’ is debatable,”} requiring refinements in numerical phrasing.

Feedback varied by background: buy-side professionals emphasized transparency in feature weighting, whereas retail reviewers focused on technical and macro signals. All commented on the lack of a neutral or hold signal, which was done to align with~\cite{Theate2021}.

Overall, results validated Prompt~4’s modular design and market narrative awareness. It outperformed earlier prompts and was selected as the global policy generation prompt for the LLM--RL hybrid in Experiment~2. 

The computational costs of Experiment~1 are summarized in Table~\ref{tab:costs}. The overall cost for a single run with each prompt was approximately \$36, increasing to about \$150 when the 
writer--judge loop was included. When accounting for additional trials and development, the cumulative cost amounted to \$345. Inference time ranged from approximately 1.5 to 2~hours per asset and prompt version.

\begin{table}[!t]
\centering
\footnotesize
\caption{Token Usage and Costs}
\label{tab:costs}
\begin{tabular}{lrrrr}
\toprule
Prompt & Mean Tokens & Mean Cost(\$) & Total Tokens & Total Cost(\$) \\
\midrule
v0 & 663 & \$0.00020 & $2.0\times 10^{6}$ & \$1.19 \\
v1 & 1,760 & \$0.00043 & $3.5\times 10^{6}$ & \$5.62 \\
v2 & 2,240 & \$0.00051 & $4.5\times 10^{6}$ & \$6.48 \\
v3 & 3,300 & \$0.00067 & $6.6\times 10^{6}$ & \$8.75 \\
v4 & 8,300 & \$0.00150 & $1.6\times 10^{7}$ & \$21.60 \\
\bottomrule
\end{tabular}
\end{table}

\subsection{Experiment 2 Results}
\label{sec:exp2a_results}

This experiment addressed Objective~2 by comparing three agent architectures: (i) the benchmark RL-only~\cite{Theate2021}, (ii) the best-performing LLM prompt from Experiment~1, and (iii) a hybrid LLM+RL agent. All agents were trained in identical environments.

To determine whether the hybrid agent outperformed the benchmark, we conducted two-sided paired $t$-tests on the SR across 25 runs for each stock. The null hypothesis $H_0$ assumed no difference in mean performance: $H_0: \mu_{\text{LLM+RL}} = \mu_{\text{RL-only}}$. All resulting $p$-values were below 0.05, indicating statistically significant improvements.

\begin{table}[t]
\centering
\footnotesize
\caption{Experiment 2 Results: Sharpe Ratio}
\label{tab:exp2_sharpe}

\begin{tabular}{lrrr}
  \toprule
  \textbf{Ticker} & \textbf{LLM+RL ($\sigma)$} & \textbf{RL-Only ($\sigma)$} & \textbf{LLM-Only} \\
  \midrule
  AAPL   & 1.70 (0.43)          & 1.42 (0.05) & \textbf{2.09} \\
  AMZN   & \textbf{1.21 (0.58)} & 0.42 (0.23) & 0.84 \\
  GOOGL  & \textbf{1.16 (0.17)} & 0.23 (0.37) & 1.12 \\
  META   & 0.46 (0.75)          & 0.15 (0.61) & \textbf{0.77} \\
  MSFT   & \textbf{1.16 (0.28)} & 0.99 (0.30) & 0.50 \\
  TSLA   & \textbf{0.92 (0.19)} & 0.62 (0.60) & 0.87 \\
  \midrule
  \textbf{Mean} & \textbf{1.10} & 0.64 & 1.03 \\
  \bottomrule
\end{tabular}
\end{table}

\vspace{1em}

\begin{table}[t]
\centering
\footnotesize
\caption{Experiment 2 Results: Maximum Drawdown}
\label{tab:exp2_mdd}

\begin{tabular}{lrrr}
  \toprule
  \textbf{Ticker} & \textbf{LLM+RL ($\sigma)$} & \textbf{RL-Only ($\sigma)$} & \textbf{LLM-Only} \\
  \midrule
  AAPL   & 0.29 (0.20)          & 0.45 (0.01) & \textbf{0.28} \\
  AMZN   & 0.26 (0.12)          & \textbf{0.19 (0.14)} & 0.34 \\
  GOOGL  & 0.28 (0.06)          & \textbf{0.25 (0.18)} & 0.35 \\
  META   & 0.35 (0.11)          & 0.45 (0.27) & \textbf{0.30} \\
  MSFT   & 0.19 (0.08)          & \textbf{0.17 (0.09)} & 0.21 \\
  TSLA   & \textbf{0.46 (0.05)} & 0.65 (0.13) & 0.59 \\
  \midrule
  \textbf{Mean} & \textbf{0.31} & 0.36 & 0.35 \\
  \bottomrule
\end{tabular}
\end{table}

\begin{figure*}[t]
    \centering
    \includegraphics[width=\textwidth]{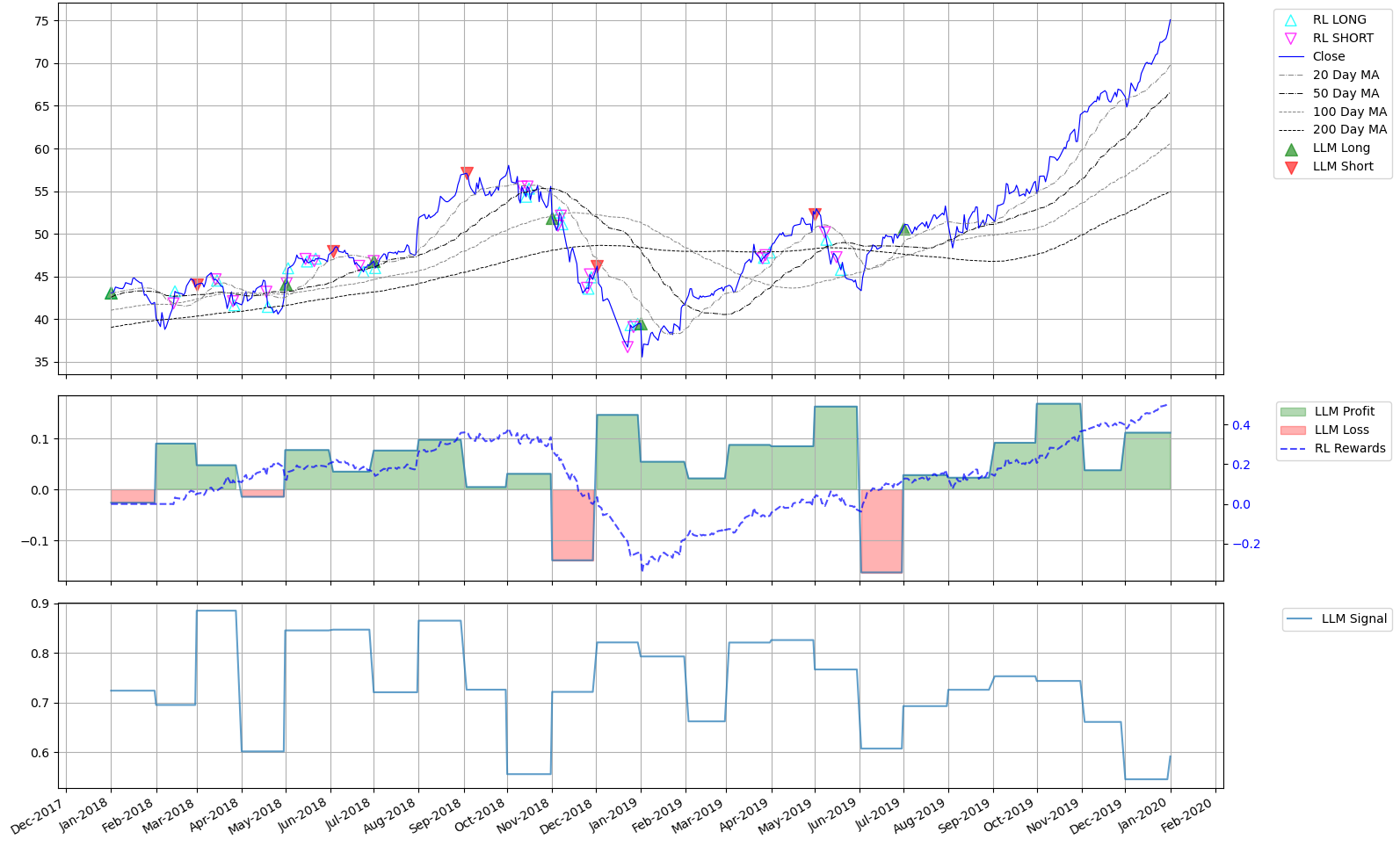}
  \caption{AAPL Performance with LLM+RL Model.}
    \label{fig:aapl_strategy_breakdown}
\end{figure*}

\begin{figure}[t]
    \centering
    \includegraphics[width=\columnwidth]{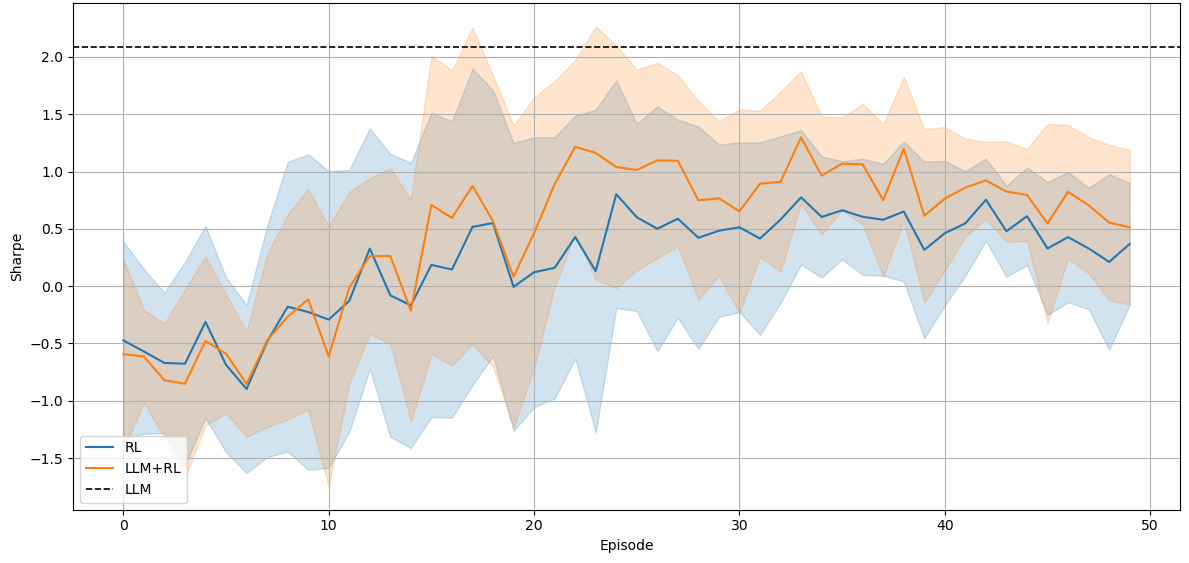}
  \caption{Training Behavior for AAPL: Sharpe Ratio.}
    \label{fig:aapl_sharpe}
\end{figure}
\begin{figure}[t]
    \centering
    \includegraphics[width=\columnwidth]{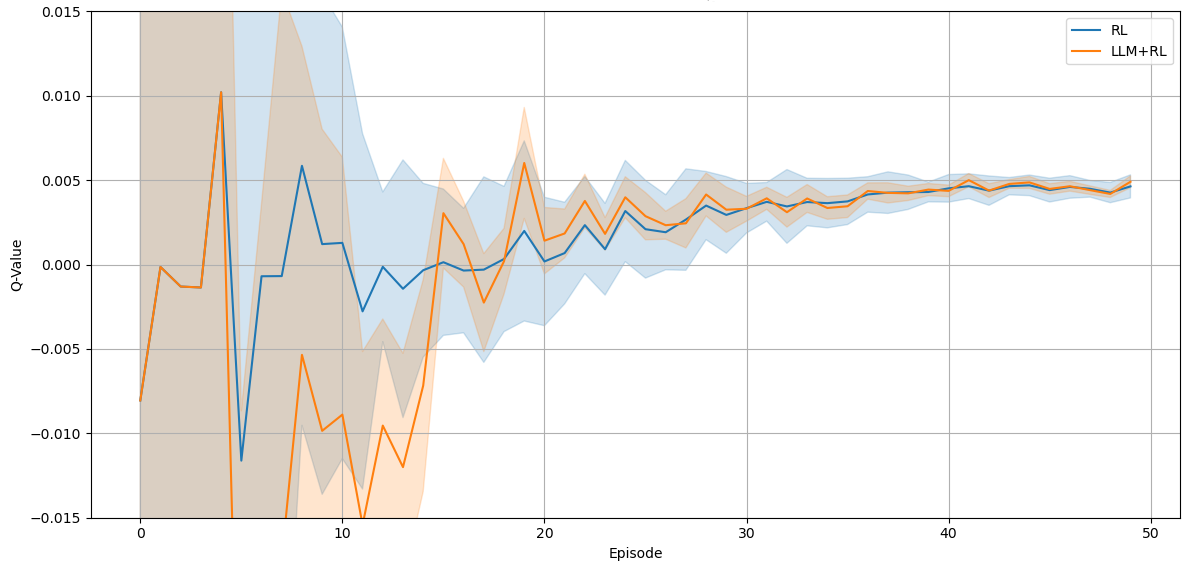}
  \caption{Training Behavior for AAPL: Q-Values for \texttt{LONG}.}
    \label{fig:aapl_ql}
\end{figure}

\begin{figure}[t]
    \centering
    \includegraphics[width=\columnwidth]{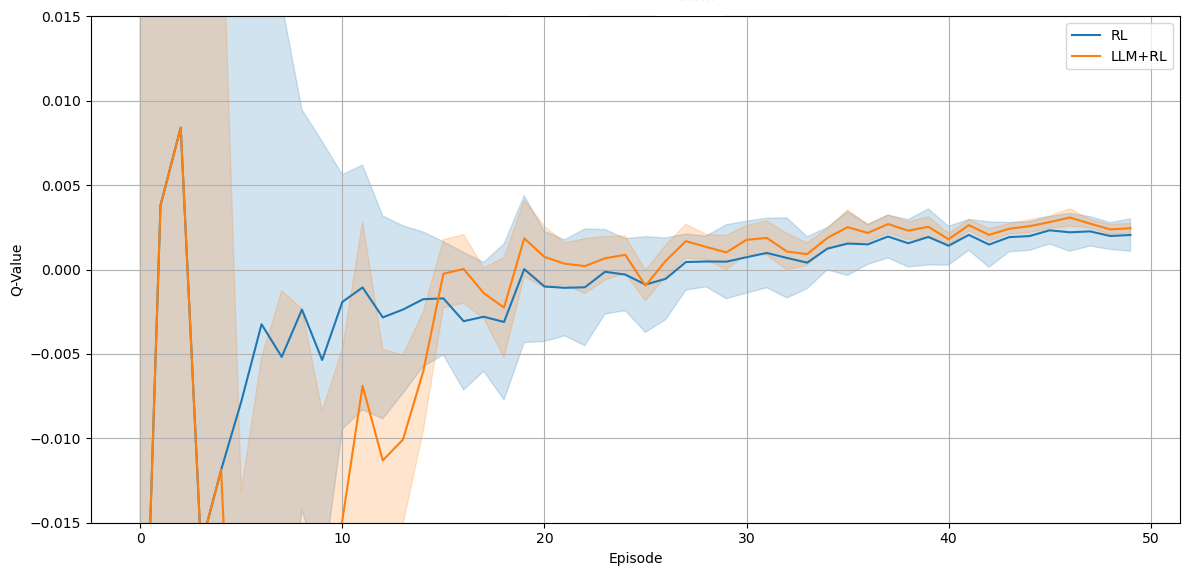}
  \caption{Training Behavior for AAPL: Q-Values for \texttt{SHORT}.}
    \label{fig:aapl_qs}
\end{figure}

Results in Table~\ref{tab:exp2_sharpe} confirm that the LLM+RL agent outperformed the RL-only baseline in four out of six assets.

\label{sec:appl_discussion}

AAPL and META did not show consistent individual outperformance. Fig.~\ref{fig:aapl_strategy_breakdown} illustrates AAPL's trading behavior during one episode. The top panel plots price, technical indicators, and trades: hollow triangles mark RL trades; filled arrows show LLM monthly guidance. The LLM issued sparse but confident signals (strength $>$ 0.6), often aligned with technical points of interest (e.g., MA interactions). In contrast, the RL agent frequently mistimed entries and exits.

From December 2018 to January 2019, the RL agent oscillated between LONG and SHORT positions with punishing results and despite receiving strong signals from the LLM. The LLM issued high-confidence guidance for a SHORT in December followed by a LONG in January, both with signal strengths exceeding 0.8. Regardless, the RL agent held a LONG position throughout the decline. 

As shown in Fig.~\ref{fig:aapl_qs}, the DDQN assigns lower Q-values to \textsc{Short} actions, indicating limited confidence. This follows from lower-bound constraints (used to cap leverage) that created an asymmetric return function by triggering buy-to-cover after price increases, reducing portfolio value and subsequent SHORT exposure. Also, the selected equity universe has positive historical drift, which raises average prices, with limited opportunity to capture SHORT returns. Together these features lower the expected return of a SHORT and discourage sustained SHORT positions~\cite{Theate2021}.

The bottom panel confirms that the LLM maintained high confidence near key inflection points, and reduced conviction when trends have persisted (possibly awaiting a reversal from its training corpus). However, the RL agent didn't fully exploit these signals due to the underlying RL architecture, which remained fixed for the purposes of this experiment.

\label{sec:rl_discussion}

Fig.~\ref{fig:aapl_sharpe} illustrates the evolution of the SR for AAPL throughout the training episodes. The hybrid LLM+RL agent (orange line) outperformed the baseline RL agent (blue line) in both mean Sharpe and stability, as reflected in the narrower shaded confidence intervals. The LLM's SR is shown for reference (black dashed line).

Figs.~\ref{fig:aapl_ql} and~\ref{fig:aapl_qs} show Q-values for \texttt{LONG} and \texttt{SHORT} actions respectively, with y-axis clipped to \([-0.03, 0.03]\) to highlight late-episode convergence. Early training was noisy for both agents. The LLM+RL agent converged faster with lower variance. Although Q-value separation rarely exceeded 0.01, the hybrid showed slightly stronger directional signals. These gains emerged without modifying the DDQN or imposing reward shaping, thus isolating the effect of the LLM's guidance. The narrow Q-range stems from the RL baseline design.

The hybrid agent did not consistently minimize MDD per stock but achieved values close to the best across agents, with the lowest overall mean (0.31). This suggests overall smoother drawdowns under uncertainty across the universe (see Tables~\ref{tab:exp2_sharpe} and~\ref{tab:exp2_mdd}).

\section{Conclusion and Future Work}

This study has explored an RL+LLM hybrid architecture for algorithmic trading, where LLMs generate guidance for RL agents to act as tactical executors.

Experiment~1 has shown that well engineered prompts improve the LLM’s performance, with Prompt~4 achieving the highest SR and lowest uncertainty. Expert evaluations confirmed the rationale of generated strategies within the domain.

Experiment~2 has demonstrated that an RL agent guided by LLM signals outperforms the RL-only baseline in four out of six stocks when evaluated by their Sharpe Ratio. While MDD was not consistently reduced, the overall drawdowns remained low on average. Importantly, the underlying RL architecture was not modified; all observed improvements stemmed from LLM guidance.

Future research should address two main directions. First, while the LLM can guide the RL, reward shaping is necessary to attain optimal results. Second, modular specialization through multiple LLM agents prompted for specific domains may enable a mixture-of-experts architecture, and lessen the risk of confabulation. 

Overall, this work presents a novel LLM+RL system that improves both return and risk outcomes. It supports modular, agentic setups where LLMs operate as trustworthy planners in financial decision making.

\section*{Supplementary Material}
Full prompt templates (strategy and analyst), labeling-heuristic pseudocode, extended dataset schema, and complete replication tables are available from the corresponding author upon request.

\section*{Acknowledgment}

We thank the expert reviewers who contributed their time and expertise to this work.

% Camera-ready placeholders (uncomment and finalize after acceptance):
% \IEEEoverridecommandlockouts
% \IEEEpubid{\makebox[\columnwidth]{978-1-XXXX-XXXX-X/25/$31.00~\copyright~2025 IEEE \hfill} \hspace{\columnsep}\makebox[\columnwidth]{ }}
% Column balancing trigger (adjust reference number as needed):
% \IEEEtriggeratref{25}
% \IEEEtriggercmd{\enlargethispage{-2in}}

\bibliographystyle{IEEEtran}
% Use IEEEabrv (abbreviated journal strings) before your database; file is in repo root.
\bibliography{IEEEabrv,references}

\begin{thebibliography}{10}
\providecommand{\url}[1]{#1}
\csname url@samestyle\endcsname
\providecommand{\newblock}{\relax}
\providecommand{\bibinfo}[2]{#2}
\providecommand{\BIBentrySTDinterwordspacing}{\spaceskip=0pt\relax}
\providecommand{\BIBentryALTinterwordstretchfactor}{4}
\providecommand{\BIBentryALTinterwordspacing}{\spaceskip=\fontdimen2\font plus
\BIBentryALTinterwordstretchfactor\fontdimen3\font minus \fontdimen4\font\relax}
\providecommand{\BIBforeignlanguage}[2]{{%
\expandafter\ifx\csname l@#1\endcsname\relax
\typeout{** WARNING: IEEEtran.bst: No hyphenation pattern has been}%
\typeout{** loaded for the language `#1'. Using the pattern for}%
\typeout{** the default language instead.}%
\else
\language=\csname l@#1\endcsname
\fi
#2}}
\providecommand{\BIBdecl}{\relax}
\BIBdecl
\renewcommand{\BIBentryALTinterwordstretchfactor}{4}

\bibitem{Chlistalla2011HFT}
\BIBentryALTinterwordspacing
M.~Chlistalla, ``High-frequency trading: Better than its reputation?'' Deutsche Bank Research, Frankfurt am Main, Germany, Tech. Rep. Research Briefing, Feb. 2011. [Online]. Available: \url{https://www.finextra.com/finextra-downloads/featuredocs/prod0000000000269468.pdf}
\BIBentrySTDinterwordspacing

\bibitem{FamaFrench2015}
E.~F. Fama and K.~R. French, ``A five-factor asset pricing model,'' \emph{Journal of Financial Economics}, vol. 116, no.~1, pp. 1--22, 2015.

\bibitem{Theate2021}
T.~Th\'{e}ate and D.~Ernst, ``An application of deep reinforcement learning to algorithmic trading,'' \emph{Expert Systems with Applications}, vol. 173, p. 114632, Jul. 2021.

\bibitem{bartram2020artificial}
S.~M. Bartram, J.~Branke, and M.~Motahari, \emph{Artificial Intelligence in Asset Management}.\hskip 1em plus 0.5em minus 0.4em\relax CFA Institute Research Foundation, 2020.

\bibitem{Sutton1998}
\BIBentryALTinterwordspacing
R.~S. Sutton and A.~G. Barto, \emph{Reinforcement Learning: An Introduction}, 2nd~ed., ser. Adaptive Computation and Machine Learning series.\hskip 1em plus 0.5em minus 0.4em\relax MIT Press, 2018. [Online]. Available: \url{http://incompleteideas.net/book/the-book-2nd.html}
\BIBentrySTDinterwordspacing

\bibitem{Liu2021FinRL}
X.-Y. Liu, H.~Yang, J.~Gao, and C.~D. Wang, ``Finrl: Deep reinforcement learning framework to automate trading in quantitative finance,'' in \emph{Proceedings of the Second ACM International Conference on AI in Finance}.\hskip 1em plus 0.5em minus 0.4em\relax ACM, Nov. 2021.

\bibitem{LopezdePrado2020BeyondEconometrics}
\BIBentryALTinterwordspacing
M.~M.~L. de~Prado, ``Beyond econometrics: A roadmap towards financial machine learning,'' \emph{Econometric Modeling: Theoretical Issues in Microeconometrics eJournal}, 2019. [Online]. Available: \url{https://api.semanticscholar.org/CorpusID:199365784}
\BIBentrySTDinterwordspacing

\bibitem{Onozo2024FinNews}
L.~Onozo, F.~Arthur, and B.~Gyires-T\'{o}th, ``Leveraging {LLMs} for financial news analysis and macroeconomic indicator nowcasting,'' \emph{IEEE Access}, 2024, early Access. Online. Accessed: Feb. 10, 2025.

\bibitem{zhang2024multimodalfoundationagentfinancial}
W.~Zhang \emph{et~al.}, ``A multimodal foundation agent for financial trading: Tool-augmented, diversified, and generalist,'' New York, NY, USA, p. 4314–4325, 2024.

\bibitem{wang2024llmfactorextractingprofitablefactors}
\BIBentryALTinterwordspacing
M.~Wang, K.~Izumi, and H.~Sakaji, ``{LLMFactor}: Extracting profitable factors through prompts for explainable stock movement prediction,'' 2024. [Online]. Available: \url{https://arxiv.org/abs/2406.10811}
\BIBentrySTDinterwordspacing

\bibitem{schulhoff2024promptreportsystematicsurvey}
\BIBentryALTinterwordspacing
S.~Schulhoff \emph{et~al.}, ``The prompt report: A systematic survey of prompting techniques,'' 2024. [Online]. Available: \url{https://arxiv.org/abs/2406.06608}
\BIBentrySTDinterwordspacing

\bibitem{Yu2024FinMem}
Y.~Yu \emph{et~al.}, ``Finmem: A performance-enhanced llm trading agent with layered memory and character design,'' in \emph{Proceedings of the AAAI Spring Symposium Series}, R.~P.~A. Petrick and C.~W. Geib, Eds.\hskip 1em plus 0.5em minus 0.4em\relax AAAI Press, Jan. 2024, pp. 595--597.

\bibitem{Arulkumaran2017}
K.~Arulkumaran, M.~P. Deisenroth, M.~Brundage, and A.~A. Bharath, ``Deep reinforcement learning: A brief survey,'' \emph{IEEE Signal Processing Magazine}, vol.~34, no.~6, pp. 26--38, 2017.

\bibitem{lopezlira2023chatgptforecaststockprice}
A.~Lopez-Lira and Y.~Tang, ``Can chatgpt forecast stock price movements? return predictability and large language models,'' 2023.

\bibitem{huang2022reasoninglargelanguagemodels}
\BIBentryALTinterwordspacing
J.~Huang and K.~C.-C. Chang, ``Towards reasoning in large language models: A survey,'' 2022. [Online]. Available: \url{https://arxiv.org/abs/2212.10403}
\BIBentrySTDinterwordspacing

\bibitem{LopezdePrado2018ML}
M.~Lopez~de Prado, ``The 10 reasons most machine learning funds fail,'' The Journal of Portfolio Management, Tech. Rep., 06 2018.

\bibitem{Lucas2024DQNSparse}
L.~Takara, A.~Santos, V.~Mariani, and L.~Coelho, ``Deep reinforcement learning applied to a sparse-reward trading environment with intraday data,'' \emph{Expert Systems with Applications}, vol. 238, p. 121897, 2024.

\bibitem{chaddha2022predictive}
A.~Chaddha and S.~Yadav, ``Examining the predictive power of moving averages in the stock market,'' \emph{Journal of Student Research}, vol.~11, no.~3, 2022.

\bibitem{wang2024quantagentseekingholygrail}
\BIBentryALTinterwordspacing
S.~Wang, H.~Yuan, L.~M. Ni, and J.~Guo, ``Quantagent: Seeking holy grail in trading by self-improving large language model,'' 2024. [Online]. Available: \url{https://arxiv.org/abs/2402.03755}
\BIBentrySTDinterwordspacing

\bibitem{lopezlira2025memorizationproblemtrustllms}
\BIBentryALTinterwordspacing
A.~Lopez-Lira, Y.~Tang, and M.~Zhu, ``The memorization problem: Can we trust {LLMs'} economic forecasts?'' 2025. [Online]. Available: \url{https://arxiv.org/abs/2504.14765}
\BIBentrySTDinterwordspacing

\bibitem{gonen-etal-2023-demystifying}
H.~Gonen, S.~Iyer, T.~Blevins, N.~Smith, and L.~Zettlemoyer, ``Demystifying prompts in language models via perplexity estimation,'' in \emph{Findings of the Association for Computational Linguistics: EMNLP 2023}, H.~Bouamor, J.~Pino, and K.~Bali, Eds.\hskip 1em plus 0.5em minus 0.4em\relax Singapore: Association for Computational Linguistics, Dec. 2023, pp. 10\,136--10\,148.

\bibitem{kaltchenko2025entropy}
\BIBentryALTinterwordspacing
A.~Kaltchenko, ``Entropy heat-mapping: Localizing {GPT}-based {OCR} errors with sliding-window shannon analysis,'' 2025. [Online]. Available: \url{https://arxiv.org/abs/2505.00746}
\BIBentrySTDinterwordspacing

\bibitem{Demajo_2020}
\BIBentryALTinterwordspacing
L.~M. Demajo, V.~Vella, and A.~Dingli, ``Explainable {AI} for interpretable credit scoring,'' 2020. [Online]. Available: \url{https://arxiv.org/abs/2012.03749}
\BIBentrySTDinterwordspacing

\bibitem{yona2024faithful}
\BIBentryALTinterwordspacing
G.~Yona, R.~Aharoni, and M.~Geva, ``Can large language models faithfully express their intrinsic uncertainty in words?'' 2024. [Online]. Available: \url{https://arxiv.org/abs/2405.16908}
\BIBentrySTDinterwordspacing

\end{thebibliography}

\appendices

\section{Strategy Prompt}
\label{app:promptv4}

The final tuned prompt from Experiment~1 and the LLM strategy generator for Experiment~2, is available in~\ref{lst:final_prompt}.

\begin{lstlisting}[language=yaml, caption={Tuned Strategy Prompt}, label={lst:final_prompt}]
User_Context:
  Last_Strategy_Used_Data:
    last_returns: "{Last_LLM_Strat_Returns}"
    last_action: "{Last_LLM_Strat_Action}"
    Rationale: |
       """{Last_LLM_Strat}"""


  Stock_Data:
    General:
      Beta: {Market_Beta}
      Classification: {classification}

    Last_Weeks_Price:
      Close: "{Close}"
      Volume: "{Volume}"

    Weekly_Past_Returns: "{Weekly_Past_Returns}"

    Historical_Volatility:
      HV_Close: "{HV_Close}"

    Implied_Volatility:
      IV_Close: "{IV_Close}"

  Fundamental_Data:
    Ratios:
      Current_Ratio: "{Current_Ratio}"
      Quick_Ratio: "{Quick_Ratio}"
      Debt_to_Equity_Ratio: "{Debt_to_Equity_Ratio}"
      PE_Ratio: "{PE_Ratio}"
    Margins:
      Gross_Margin: "{Gross_Margin}"
      Operating_Margin: "{Operating_Margin}"
      Net_Profit_Margin: "{Net_Profit_Margin}"
    Growth Metrics:
      EPS_YoY: "{EPS_YoY_Growth}"
      Net_Income_YoY: "{Net_Income_YoY_Growth}"
      Free_Cash_Flow_YoY: "{Free_Cash_Flow_Per_Share_YoY_Growth}"

  Technical_Analysis:
    Moving_Averages:
        20MA: "{20MA}"
        50MA: "{50MA}"
        200MA: "{200MA}"
    MA_Slopes:
        20MA_Slope: "{20MA_Slope}"
        50MA_Slope: "{50MA_Slope}"
        100MA_Slope: "{100MA_Slope}"
        200MA_Slope: "{200MA_Slope}"
    MACD:
        Value: "{MACD}"
        Signal_Line: "{Signal_Line}"
        MACD_Strength: {MACD_Strength}
    RSI:
        Value: "{RSI}"
    ATR: "{ATR}"

  Macro_Data:
    Macro_Indices:
      SPX:
        Close: "{SPX_Close}"
        Close_20MA: "{SPX_Close_MA}"
        Close_Slope: "{SPX_Close_Slope}"
      VIX:
        Close: "{VIX_Close}"
        Close_20MA: "{VIX_Close_MA}"
        Close_Slope: "{VIX_Close_Slope}"
    Economic_Data:
      GDP_QoQ: "{GDP_QoQ}"
      PMI: "{PMI}"
      Consumer_Confidence_QoQ: "{Consumer_Confidence_QoQ}"
      M2_Money_Supply_QoQ: "{M2_Money_Supply_QoQ}"
      PPI_YoY: "{PPI_YoY}"
      Treasury_Yields_YoY: "{Treasury_Yields_YoY}"

  Options_Data:
    Put_IV_Skews:
      OTM_Skew: "{OTM_Skew}"
      ATM_Skew: "{ATM_Skew}"
      ITM_Skew: "{ITM_Skew}"
    20Day_Moving_Averages:
      OTM_Skew_MA: "{MA_OTM_Skew}"
      ATM_Skew_MA: "{MA_ATM_Skew}"
      ITM_Skew_MA: "{MA_ITM_Skew}"

  News_Sentiment: {news_sentiment}
  News_Impact_Score: {news_impact_score}

System_Context(System):
  Persona: {persona}
  Portfolio_Objectives: {portfolio_objectives}
  Instructions: |
    Develop a LONG or SHORT trading strategy for a single stock only for the next Month that aligns with the `portfolio_objectives`. Follow these guidelines:

    1. Stock Analysis:
       - Evaluate price trends: Compare the Close price against 20MA, 50MA, and 200MA to assess momentum or reversals.
       - Analyze returns: Use Weekly Past Returns to validate trend sustainability.
       - Contextualize volatility: Align `HV_Close` and `HV_High` with recent price action for trend validation.
       - Incorporate beta: Use `beta` to gauge sensitivity to market movements.

       - ICL Example: "Close price above 20MA and 50MA with steep 20MA slope signals bullish momentum. Weekly returns confirm a sustainable uptrend."

    2. Technical Analysis:
       - Use RSI: Identify momentum signals (>70 overbought; <30 oversold) and divergences for reversals.
       - Validate with `MACD`: Use crossovers of `MACD.Value` and `Signal_Line`, and `MACD_Strength` for directional confidence.
       - Leverage `RSI.value` divergences, and steep `Moving_Averages` slopes. Or focus on stable `Moving_Averages` patterns on stable historical volatility `HV_Close`.
       - ICL Example: "RSI at 65, a positive MACD crossover indicate bullish momentum."

    3. Fundamental Analysis:
       - Evaluate growth metrics: Use `EPS_YoY`, `Net_Income_YoY`, and `Free_Cash_Flow_YoY`for profitability and sustainability.
       - Prioritize ratios: Low `Debt_to_Equity_Ratio` and `Current_Ratio` reflect financial stability.
       - Focus on aggressive `Growth Metrics` and earnings news.
       - ICL Example: "EPS YoY growth of 25% and low Debt-to-Equity ratio of 0.5 support strong financial health, aligning with a LONG strategy."

    4. Macro Analysis:
      - Align with market sentiment across `Macro_Data`:
        - "SPX_Close_Slope > 0 && VIX_Close_Slope < 0": Bullish (Risk-On)
        - "SPX_Close_Slope < 0 && VIX_Close_Slope > 0": Bearish (Risk-Off)
      - Validate with `Economic_Data`:
        - "GDP_QoQ > 0 && `PMI` > 50" leads to Economic Expansion
        - "`Treasury_Yields_YoY` < 0" Signals Recession Risk, especially if already mentioned in `Rationale`.

      - ICL Examples:
        - "`SPX_Close_Slope` > 0 && `VIX_Close_Slope` < 0 We have Market Confidence"
        - "`GDP_QoQ` Falling && `PMI` < 50 We have an Economic Slowdown."

    5. Options Analysis:
      - Compare `OTM_Skew`, `ATM_Skew`, and `ITM_Skew` IV Skews: Assess differences to gauge market sentiment and directional bias using their `20Day_Moving_Averages`.
      - Leverage IV spikes to capitalize on speculative directional trades.
      - Example: "Rising `ATM_Skew_MA` > 0, market pricing up move, with stable HV supports a LONG position, as it indicates growing upside expectations without excessive fear."

    6. News Analysis:
      - Use `News_Sentiment` and `News_Impact_Score` (1-3).
      - Only strong directional news (score = 3) should override other signals.
      - Medium news (score = 2) supports but does not lead.
      - Always check if news contradicts macro or technical trend.

    7. Performance Reflection and Strategic Adaptation:
      - If `Last_Strategy_Used_Data` is available:
            - Assess the outcome of the previous strategy by examining `last_returns` and the chosen `last_action`.
            - Determine if the result aligns with the expectations outlined in the previous `Rationale`.
            - Identify if the direction (LONG or SHORT) led to desirable or undesirable outcomes.
            - You must NOT reuse or copy the previous `Rationale`. It is only context for reflection.
            - Summarize in 1-2 sentences whether the previous strategy performed as expected.
            - Example: "The previous LONG strategy yielded positive returns, confirming the bullish setup based on RSI and moving averages."
            - Do NOT include language or phrasing from the previous rationale.
      - Confidence assignment:
          - Assign a Likert score (1 to 3) to your  `action_confidence`:
              - 1: Low confidence; contradictory or weak alignment across features.
              - 2: Moderate confidence; partial alignment with moderate evidence.
              - 3: High confidence; strong convergence across key features.
      - Feature Attribution:
          - Rank the importance of each major feature used in your current rationale using a Likert scale (1 to 3):
              - 1: Minimal contribution; not required for the decision.
              - 2: Moderate contribution; relevant but not critical.
              - 3: High contribution; pivotal to the trading decision.

Output:
  action: Str. LONG or SHORT.
  action_confidence: int. Likert scale (1-3) confidence in the proposed `action`, adjusted based on prior strategy outcome if `Last_Strategy_Used_Data` is available.
  explanation: >
    A concise rationale (max 350 words) justifying the proposed `action`.
    Include:
      - The top 5 weighted features used in the decision, each labeled with its Likert importance (1-3).
        (e.g., "Stock_Data.Price.Close, Weight 3, Technical_Analysis.RSI.Value, Weight 1, Options_Data.ATM_Skew, Weight 2")
      - A reflective assessment of `Last_Strategy_Used_Data`, including whether the past `action` was successful and was it maintained given prior `Rationale`.
  features_used:
    - feature: the features used from the prompt's context.
      direction: LONG, SHORT, or NEUTRAL
      weight: A Likert score (1 to 3) described in Feature Attribution.
\end{lstlisting}

\section{Analyst Prompt}
\label{app:analyst_prompts}

The Analyst prompt used in Experiment~1 is presented in Listing~\ref{lst:analyst_prompt}, adapted from~\cite{wang2024llmfactorextractingprofitablefactors}. News corpora were anonymized prior to prompting.

\begin{lstlisting}[language=yaml, caption={Analyst Prompt}, label={lst:analyst_prompt}]
User_Context:
  Monthly_News_Articles_List: |
    "{articles_list}"

System_Context:
  Persona: Financial Market Analyst
  Instructions: |
    Extract the `Top 3` news factors influencing stock price movements from the `Monthly_News_Articles_List`. Follow these steps:

    1. Rank the news by relevance to stock price movements:
       - Prioritize news related to significant financial or market impacts (e.g., acquisitions, partnerships, guidance revisions).
       - Weigh industry trends, macroeconomic influences, and analyst ratings based on their expected effect on the company valuation.
       - News with broad or long-term implications ranks higher.

    2. Summarize content into key factors and corporate events affecting stock prices, using concise language and causal relationships.

    3. For each factor, assign:
       - `Sentiment`: +1 for positive, -1 for negative, 0 for neutral or mixed
       - `Market_Impact_Score`: Likert scale from 1 to 3, where:
         - 1 = minimal relevance
         - 2 = moderate influence
         - 3 = high impact driver

    Examples of factors influencing stock prices include:
      - Strategic partnerships or competitor activity.
      - Industry trends or macroeconomic influences.
      - Product launches or market expansions.
      - Analyst ratings, significant stock price moves, or expectations.
      - Corporate events: guidance revisions, acquisitions, contracts, splits, repurchases, dividends.

    Example:
      'A major tech company partners with a leading automotive firm for EV battery innovation. Analysts predict this could boost revenues significantly.'

      Ranked Factors:
        1. factor: Strategic partnership in EV battery technology expected to increase revenue.
           sentiment: +1
           market_impact: 3
        2. factor: Positive sentiment driven by projected long-term gains.
           sentiment: +1
           market_impact: 2
        3. factor: Growing demand for EV technology anticipated to support future earnings.
           sentiment: +1
           market_impact: 2

Output:
  factors:
    - factor: str. Summary of the news item. Max 70 words.
    - sentiment: int. One of Positive +1, Negative -1, or Neutral 0
    - market_impact: int. Likert scale 1 to 3
\end{lstlisting}

\section{Algorithms}

The labeling algorithm emulates expert trading behavior by deliberately leveraging future return information to assign proxy trade actions in hindsight. This approach offers a cost-effective and scalable addition to manual annotation, capturing the general direction an informed trader might take. These synthetic labels are then provided to the LLM, along with a smaller set of HITL annotated examples.

\begin{algorithm}[t]
\caption{Expert Trade Heuristic}
\label{alg:expert_action}
\KwData{Time-indexed price series}
\KwResult{Trade action: LONG (1) or SHORT (0)}

\ForEach{date \( t \) in dataset}{
    \( P_t \gets \text{Close}(t) \)\;
    \( r^{(10)} \gets \frac{P_{t+10}}{P_t} - 1 \), 
    \( r^{(20)} \gets \frac{P_{t+20}}{P_t} - 1 \)\;
    \( r^{\text{weighted}} \gets 0.4 \cdot r^{(10)} + 0.6 \cdot r^{(20)} \)\;
    \eIf{\( r^{\text{weighted}} >= 0 \)}{
        \textbf{Action} $\gets$ LONG \quad (\texttt{Trade\_Action} $= 1$)\;
    }{
        \textbf{Action} $\gets$ SHORT \quad (\texttt{Trade\_Action} $= 0$)\;
    }
}
\end{algorithm}

\section{Dataset}
\label{app:dataset}

\subsection*{Market Data} This market data (\(\mathcal{S}_{\text{mk}}\)) included OHLCV price series as well as macro-level indicators and forward-looking sentiment signals. Specifically, it comprised:
\begin{itemize}
    \item Daily returns of the S\&P 500 Index (SPX) and NASDAQ-100 Index (NDX). These are market and sector indices,
    \item Implied Volatility (IV) and Historical Volatility (HV) metrics, derived from the stock's derivatives,
    \item The CBOE Volatility Index (VIX) as a proxy for market fear and option market expectations,
    \item \textit{Weekly Past Returns}, which record the percentage change over the past four weekly intervals. The four-week span was selected empirically to align with the model’s monthly strategy generation frequency.
\end{itemize}
These features help in modeling short-term market dynamics.

\subsection*{Fundamental Data}

Fundamental data (\(\mathcal{S}_{\text{fund}}\)) has firm-level fundamentals and macroeconomic indicators. Macroeconomic variables provided contextual narrative for interpreting observed signals, and supporting regime identification~\cite{zhang2024multimodalfoundationagentfinancial, Onozo2024FinNews}. This set covered:
\begin{itemize}
    \item \textbf{Liquidity ratios:} Current Ratio, Quick Ratio;
    \item \textbf{Leverage and coverage:} Debt-to-Equity, Interest Coverage;
    \item \textbf{Profitability metrics:} Gross Margin, Operating Margin, Return on Equity (ROE), Return on Assets (ROA);
    \item \textbf{Valuation:} Price-to-Earnings (P/E), Price-to-Book (P/B), Enterprise Value (EV), and Earnings Before Interest, Taxes, Depreciation, and Amortization (EBITDA).
    \item \textbf{Growth:} Revenue and Earnings Growth;
    \item \textbf{Macroeconomic indicators:} Gross Domestic Product (GDP), Purchasing Managers’ Index (PMI), Producer Price Index (PPI), Consumer Confidence Index (CCI), U.S. 10-Year Treasury Yield, and the 10Y–2Y yield curve slope.
\end{itemize}

To enhance temporal abstraction, all variables were computed as quarter-over-quarter (QoQ) or year-over-year (YoY) percentage changes. It is critical to take first-order dynamics as LLMs can recall absolute numbers for economic details, allowing look-ahead bias in the backtests \cite{lopezlira2025memorizationproblemtrustllms}.

\subsection*{Analytics}

Technical indicators (\(\mathcal{S}_{\text{an}}\)) were computed over rolling 20-day windows using the open-source TA-Lib\footnote{\url{https://ta-lib.org/}} library. These features include:
\begin{itemize}
    \item Simple Moving Averages (SMA) over 20, 50, 100, 200 trading-day horizons,
    \item Relative Strength Index (RSI),
    \item Average True Range (ATR) for volatility,
    \item Moving Average Convergence Divergence (MACD) with its signal line and derived strength,
    \item Volume-Weighted Average Price (VWAP) as a reference anchor for intraday valuations.
\end{itemize}
Each indicator was extended with slope and z-score to assist the LLM in capturing directional shifts and the statistical significance of deviations. These technical indicators are widely used in trading practice and academic research~\cite{chaddha2022predictive}.

\subsection*{Alternative Data}

Structured representations of financial news headlines ($\mathcal{S}_{\text{alt}}$) were extracted using a large language model (LLM), which anonymized and synthesized the content into latent factors. Following the LLMFactor methodology~\cite{wang2024llmfactorextractingprofitablefactors}, each news item was distilled into 2–5 interpretable factors, capturing macroeconomic and firm-specific signals.

To mitigate memorization and data leakage risks, named entities and dates were anonymized (e.g., “Tesla” becomes “the Company”).

\section{Replicated Benchmark Metrics}
\label{app:bm_metrics}

We report the replicated benchmark metrics in Appendix~\ref{app:bm_metrics} for the assets used in~\cite{Theate2021}. We include the mean SR and MDD, each averaged across 25 runs with standard deviation $\sigma$.

For the SR, we conduct a two-sided one-sample $t$-test to assess whether the metric is significantly different from the published value. The null hypothesis $H_0$ assumes equivalence, i.e. $H_0: \mu_{\text{SR}} = \text{SR}_{\text{paper}}$. 

Since this is a replication test, failing to reject $H_0$ indicates successful replication. $p$-values are computed only for SR; other metrics are reported without significance testing. 

All assets have been successfully replicated within acceptable bounds, with exceptions highlighted in bold. Notably, GOOGL, one of the stocks included in our test environment, exhibited a statistically significant deviation from the original benchmark, with a $p$-value below 0.05.

\begin{table}[t]
\footnotesize
\centering
\begin{tabular}{lrrr}
\toprule
\textbf{Instrument} & \textbf{Paper SR} & \textbf{SR (±$\sigma$) [\textbf{$p$-value}]} & \textbf{MDD (±$\sigma$)} \\  
\midrule
AB InBev & 0.187 & \textbf{1.21 (0.30) [0.00]} & 0.18 (0.08) \\ 
Alibaba & 0.021 & \textbf{0.06 (0.02) [0.00]} & 0.09 (0.01) \\
Amazon & 0.419 & 0.39 (0.45) [0.85] & 0.30 (0.09) \\
Apple & 1.424 & 1.19 (0.55) [0.22] & 0.29 (0.09) \\ 
Baidu & 0.080 & \textbf{0.20 (0.17) [0.00]} & 0.36 (0.09) \\ 
CCB & 0.202 & \textbf{0.33 (0.25) [0.04]} & 0.24 (0.14) \\ 
Coca Cola & 1.068 & 1.07 (0.53) [0.50] & 0.25 (0.04) \\
Dow Jones & 0.684 & 0.70 (0.30) [0.91] & 0.25 (0.05) \\ 
ExxonMobil & 0.098 & 0.10 (0.35) [0.91] & 0.34 (0.08) \\
FTSE 100 & 0.103 & \textbf{0.50 (0.23) [0.00]} & 0.31 (0.08) \\
Google & 0.227 & \textbf{-0.54 (0.59) [0.00]} & 0.43 (0.13) \\
HSBC & 0.011 & \textbf{0.38 (0.17) [0.00]} & 0.29 (0.05) \\
JPMorgan Chase & 0.722 & 0.72 (0.31) [0.98] & 0.26 (0.06) \\
Kirin & 0.852 & 0.85 (0.42) [0.99] & 0.39 (0.07) \\
Meta & 0.151 & \textbf{0.63 (0.61) [0.01]} & 0.45 (0.27) \\
Microsoft & 0.987 & 0.70 (1.00) [0.38] & 0.28 (0.16) \\
NASDAQ 100 & 0.845 & 0.85 (0.35) [1.00] & 0.16 (0.05) \\
Nikkei 225 & 0.019 & \textbf{0.26 (0.29) [0.02]} & 0.29 (0.07) \\
Nokia & -0.094 & \textbf{0.07 (0.24) [0.00]} & 0.57 (0.15) \\
PetroChina & 0.156 & 0.22 (0.29) [0.29] & 0.67 (0.00) \\
Philips & 0.675 & \textbf{1.40 (0.50) [0.00]} & 0.25 (0.03) \\
S\&P 500 & 0.834 & 0.83 (0.25) [1.00] & 0.14 (0.04) \\
Shell & 0.425 & 0.42 (0.37) [0.95] & 0.51 (0.05) \\
Siemens & 0.426 & 0.39 (0.23) [0.43] & 0.26 (0.12) \\
Sony & 0.424 & 0.42 (0.36) [0.97] & 0.16 (0.04) \\
Tesla & 0.621 & 0.48 (0.41) [0.29] & 0.52 (0.09) \\
Tencent & -0.198 & -0.19 (0.33) [0.98] & 0.10 (0.09) \\
Toyota & 0.304 & 0.36 (0.27) [0.37] & 0.45 (0.10) \\
Volkswagen & 0.216 & \textbf{0.45 (0.18) [0.00]} & 0.48 (0.09) \\
\bottomrule
\end{tabular}
\vspace{2pt}
\caption{Replication Metrics for \cite{Theate2021}}
\label{tab:papermetrics}
\end{table}

\end{document}